\documentclass[10pt,twocolumn,letterpaper]{article}

\usepackage[pagenumbers]{cvpr}
\usepackage{booktabs}
\usepackage{multirow}
\usepackage{xcolor}
\usepackage{colortbl}

\usepackage{cuted}
\usepackage{caption}
\definecolor{cvprblue}{rgb}{0.21,0.49,0.74}
\usepackage[pagebackref,breaklinks,colorlinks,allcolors=cvprblue]{hyperref}

\def\confName{CVPR}
\def\confYear{2027}

\title{ALIVE: Bring Foundation Features to Life with Events}

\author{
    Rong Zou\textsuperscript{1} \quad
    Haiqing Hao\textsuperscript{2} \quad
    Davide Scaramuzza\textsuperscript{1} \\
    \textsuperscript{1}Robotics and Perception Group, University of Zurich \quad
    \textsuperscript{2}Tsinghua University\\[-30pt]
}

\begin{document}
\maketitle
\begin{strip}
    \centering
    \includegraphics[width=\textwidth]{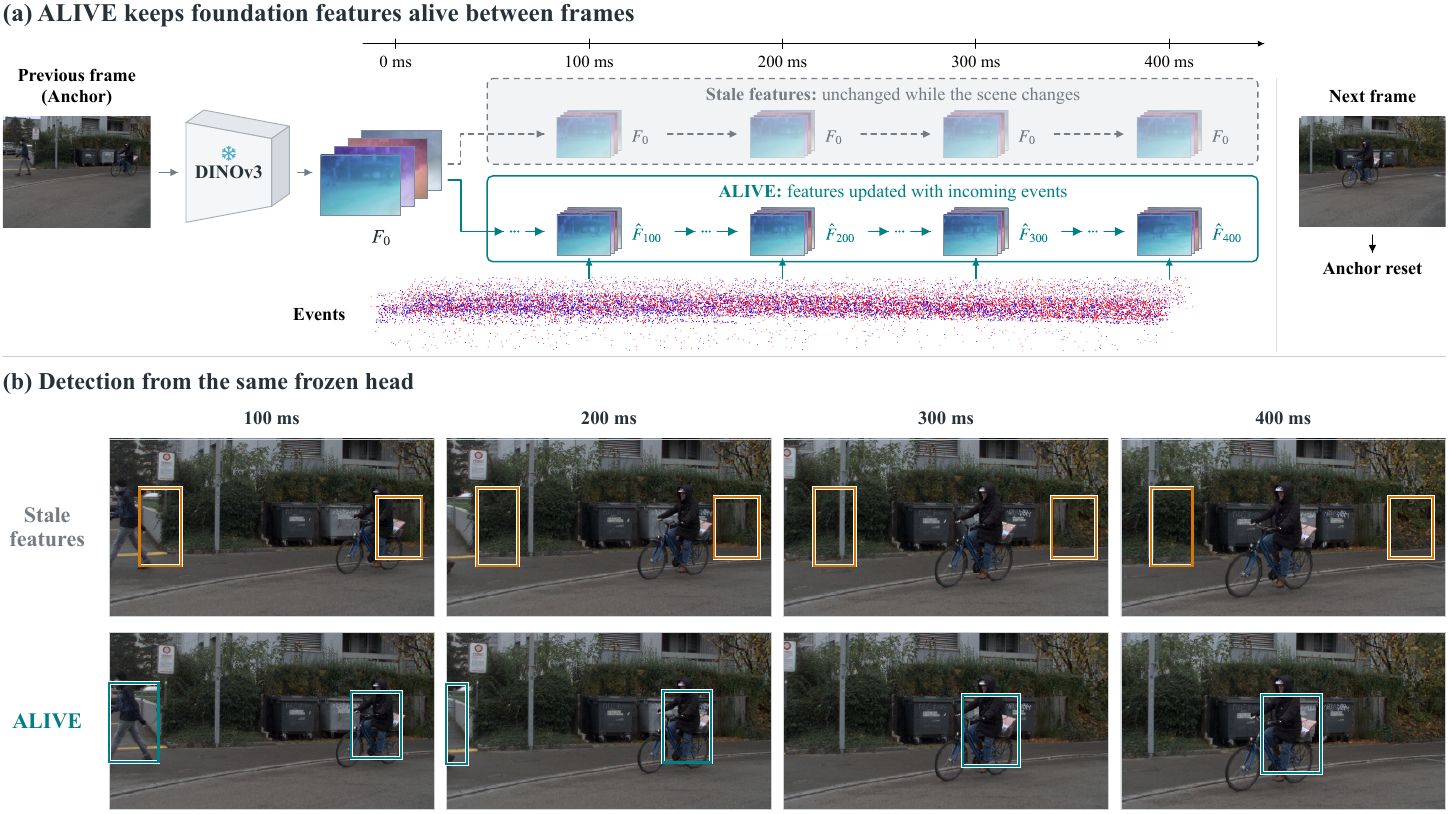}
    \captionsetup{hypcap=false}
    \captionof{figure}{
        \textbf{ALIVE keeps foundation features alive between frames.} (a) A frozen DINOv3 encodes the latest frame into multi-layer features. Before the next frame arrives, incoming events recurrently update these foundation features, while stale features remain unchanged. (b) Detections from the same frozen head during 400\,ms of continuous updates, shown at 100\,ms intervals. Both branches start from identical anchor features. \textbf{Intermediate RGB frames serve only as visualization backgrounds and are not inputs to ALIVE.}
    }
    \captionsetup{hypcap=true}
    \label{fig:teaser}
\end{strip}

\begin{abstract}
Visual foundation models provide powerful representations for diverse tasks, but their features become stale as visual content changes between image observations. Refreshing them requires new frames and repeated runs of a computationally expensive image encoder.
We introduce \textbf{ALIVE}, an event-based method that incrementally updates native foundation model features between image observations. Given an anchor image, ALIVE uses subsequent events to transport and correct existing multi-layer features extracted by a frozen DINOv3 encoder. Its lightweight updater is trained end-to-end using only synthetic data, without downstream task supervision, and transfers directly to real-world event streams without any adaptation.
Across three real-world datasets and five task--dataset settings spanning object detection, semantic segmentation, and depth estimation, ALIVE outperforms all evaluated event-based methods in every metric. On DSEC, when the anchor frame is 200\,ms old, ALIVE improves over anchor-image predictions by 30.9\,mAP in detection, 11.6\,mIoU in segmentation, and 15.9\% in depth AbsRel. Compared with image re-encoding, ALIVE reduces FLOPs by 82.4\% and latency by a factor of 2.5.
Code will be released.
\end{abstract}

\section{Introduction}
\label{sec:intro}

Visual foundation models (VFMs) provide general-purpose features that a single frozen backbone can share across many tasks through lightweight heads~\cite{radford2021learning,dinov2,kirillov2023segment,dinov3}.
However, these features are computed from a single image and describe the scene only at the moment it was captured.
Between frames, they stay fixed while the scene keeps changing, and every prediction derived from them grows stale.
This staleness limits applications that must respond quickly, such as autonomous driving and robotics~\cite{gehrig2024lowlatency,li2020streaming}.

With a conventional camera, one must choose between two unsatisfactory options.
At a low frame rate, predictions remain stale until the next image arrives, and at a high frame rate, the VFM must run on every image at a high computational cost.
Event cameras offer a way out of this dilemma.
They report per-pixel brightness changes asynchronously with microsecond resolution~\cite{gallego2022survey}.
Paired with a frame camera, either on the same chip~\cite{brandli2014davis,alpsentek2023hvs,kodama2023hvs,guo2023hvs} or as a separate sensor~\cite{dsec,zou2026event,zou2023seeing}, they observe precisely the interval in which VFM features go stale.

The question then becomes how to use events to keep perception up to date.
Most event-based methods train dedicated networks with task labels, \eg, detection~\cite{perot2020learning,gehrig2023rvt,hao2026low}, segmentation~\cite{alonso2019evsegnet,ess}, or depth estimation~\cite{ramnet2021}.
Event-guided feature propagation~\cite{lifrseg,lifrv2} keeps the features of task-specific networks up to date between frames but is likewise trained with task supervision.
Each new task thus requires new labels and new training.
Task-agnostic alternatives avoid task labels, but each falls short in efficiency or accuracy.
Reconstructing images, from events alone~\cite{e2vid,hypere2vid} or from events and frames~\cite{vfpsie}, and encoding them with the VFM brings back the cost of running the VFM on every image.
Warping features with event-based optical flow~\cite{eraft,edcflow} only moves existing content and cannot account for changes that motion does not explain.
Mapping events directly into the feature space of a VFM~\cite{scaleevent,realm2026} discards the rich visual information already captured by the latest frame.
What remains missing is a task-agnostic update that builds on the latest frame, accounts for changes beyond motion, and costs only a fraction of re-encoding.

\begin{figure*}[t]
    \centering
    \includegraphics[width=\textwidth]{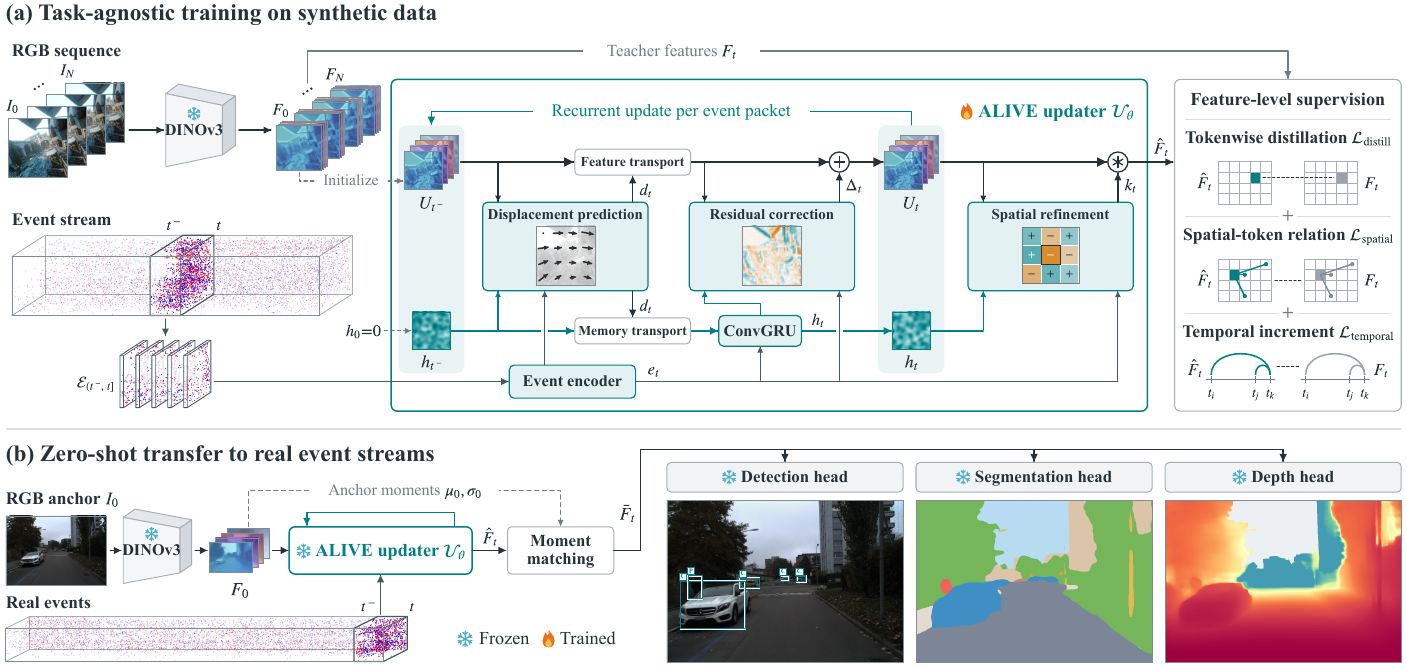}
    \caption{
        \textbf{Overview of ALIVE.}
        (a) Training uses synthetic RGB--event sequences.
        A frozen DINOv3 encodes every RGB frame: $F_0$ initializes the feature state (with memory $h_0=0$), and later frames provide teacher features $F_t$.
        For each event packet $\mathcal{E}_{(t^-,t]}$, the ALIVE updater $\mathcal{U}_\theta$ predicts a displacement $d_t$ that transports both the features and the memory, a residual correction $\Delta_t$ that yields $U_t$, and a signed, zero-sum $3\times3$ kernel $k_t$ per token that refines $U_t$ into $\hat{F}_t$.
        The ConvGRU memory $h_t$ conditions the residual and the refinement.
        All modules are trained jointly with feature-level losses only, without task labels or task heads.
        (b) The trained updater is applied without adaptation to real event streams.
        Moment matching aligns the per-channel statistics of $\hat{F}_t$ with the anchor moments to produce the updated features $\bar{F}_t$,
         and frozen task heads trained on DINOv3 features of real RGB images read $\bar{F}_t$ directly.
        Predictions are shown 200\,ms after the anchor on DSEC, and detections are drawn on the RGB frame at that time for visualization only (C: car, P: person).
    }
    \label{fig:method}
\end{figure*}

Our key insight is that such an update should act on the features of the frozen VFM and can be supervised by the VFM itself.
These features are the common input of all heads, and updating them refreshes all predictions at once.
Encoding a later frame with the same VFM yields the features that the update should produce.
The update can therefore be learned without any task label.
Based on this insight, we propose \textbf{ALIVE}, a lightweight event-driven updater that keeps the features of a frozen VFM up to date between frames (\cref{fig:teaser}).
We instantiate it with DINOv3~\cite{dinov3}.
Starting from the multi-layer features of the latest frame, ALIVE updates them with each incoming packet of events (\cref{fig:method}): it moves existing content along the motion observed by the events and corrects the changes that motion cannot explain.
At inference, parameter-free moment matching aligns the updated features with the statistics of the latest frame.

We train ALIVE solely on synthetic RGB--event sequences~\cite{tartanair,tartanairv2} and apply it zero-shot to three real-world datasets, DSEC, DDD17, and CoSEC~\cite{dsec,ddd17,cosec}, which cover object detection, semantic segmentation, and metric depth estimation in five task--dataset settings.
In each setting, a frozen head trained only on DINOv3 features of real RGB images reads the updated features.
ALIVE outperforms all compared event-based methods in all five settings.
On DSEC, when the latest frame is 200\,ms old, it improves over reusing that frame's features by 30.9\,mAP in detection, 11.6\,mIoU in segmentation, and 15.9\% in depth AbsRel (\cref{fig:accuracy_efficiency}).
ALIVE has only 2.5M parameters, and each update requires 82.4\% fewer FLOPs than re-encoding a frame with DINOv3 and runs $2.5\times$ faster, which makes it the only evaluated method fast enough to update at 200\,Hz.

In summary, our contributions are:
\begin{itemize}
    \item We propose to keep the features of a frozen VFM up to date with events between frames, which refreshes all downstream heads at once and lets the VFM itself supervise the update without task labels.
    \item We present ALIVE, a lightweight recurrent updater that transports and corrects multi-layer VFM features with events, together with a parameter-free moment matching that aligns its output with the latest frame.
    \item We show that a single ALIVE updater trained only on synthetic data transfers to real data without any adaptation, outperforms all compared methods in all five settings, and is the only one fast enough to update at 200\,Hz.
\end{itemize}
\vspace{-5pt}

\section{Related Work}
\label{sec:related}

\paragraph{Foundation Models for Event Vision.}
The scarcity of labeled event data has motivated transferring knowledge from VFMs to event cameras, for open-world recognition~\cite{eventbind2024}, language-grounded video object segmentation with Segment Anything~\cite{fedorchenko2026event,kirillov2023segment}, and depth estimation~\cite{depthanyevent2025,eventdam2025}.
Beyond individual tasks, ScaleEvent~\cite{scaleevent} and REALM~\cite{realm2026} train event encoders to reproduce dense VFM features.
These approaches recover rich representations from events alone, although events miss absolute intensity and anything that does not move relative to the camera.
Most of them also need an encoder as large as the VFM for every output.
We instead combine the strengths of both modalities: a frozen VFM extracts rich features from the latest frame, and a lightweight network updates them with new events.

\paragraph{Event-Driven High-Frequency Perception.}
Events can update visual representations between frames.
One line of work recovers images between frames, either from events alone~\cite{e2vid,hypere2vid} or by updating the latest frame with the subsequent events through direct integration~\cite{brandli2014decompression} or a learned network~\cite{vfpsie}.
Extracting VFM features from these images, however, requires a full VFM forward pass for every update.
Event-based optical flow~\cite{eraft,edcflow} can instead warp existing features, but warping only relocates them and cannot capture changes beyond motion.
ALIVE overcomes both limitations by transporting and correcting VFM features directly, at a fraction of the cost of VFM inference.

\paragraph{Event-Guided Feature Propagation.}
Closer to our work, LiFR-Seg~\cite{lifrseg} propagates segmentation features from an RGB keyframe with event-guided motion, and the concurrent LiFR-v2~\cite{lifrv2} adds event-guided feature completion and historical retrieval for segmentation and depth estimation.
Both rely on task supervision, and LiFR-v2 additionally requires a separate feature distillation stage before task-specific training.
ALIVE instead learns task-agnostic updates of frozen VFM features in a single stage, supervised only by the VFM features of subsequent frames and without modifying the VFM or the heads.
We show that a single updater trained only on synthetic data serves all evaluated tasks and real-world datasets without adaptation.

\section{Method}
\label{sec:method}
\begin{figure*}[t]
    \centering
    \includegraphics[width=\textwidth]{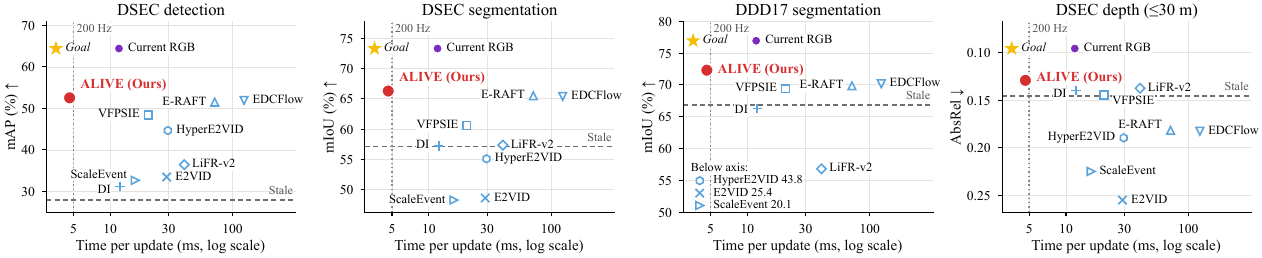}\\[3pt]
    \includegraphics[width=\textwidth]{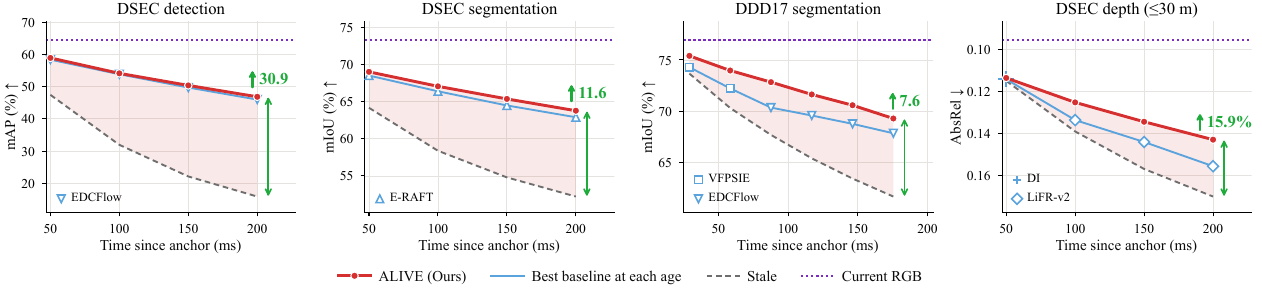}
    \caption{
        \textbf{Accuracy versus update cost (top) and versus anchor age (bottom).}
        Each column is one task--dataset setting, evaluated with a frozen task head shared by all methods.
        Top: accuracy pooled over all anchor ages against the time per feature update.
        The dashed line marks the stale anchor features, the dotted line marks the 5\,ms limit for updates at 200\,Hz, and the star marks the goal of reaching the accuracy of Current RGB within this limit.
        Current RGB encodes the target frame itself, which is unavailable between frames.
        Bottom: accuracy at each anchor age.
        The blue line connects the best baseline at each age, and its markers identify the method as in the top row.
        Arrows give the gain of ALIVE over the stale features at the longest age (for AbsRel, the relative decrease).
    }
    \label{fig:accuracy_efficiency}
\end{figure*}

\subsection{Problem Setup}
\label{sec:setup}

A frozen VFM $\Phi$ encodes the latest frame $I_0$, the anchor, into multi-layer features $F_0=\Phi(I_0)=\{F_0^\ell\}_{\ell=1}^{L}$ on its token grid, where each token corresponds to an image patch.
As the scene changes, an event camera records the resulting brightness changes, which we divide into packets $\mathcal{E}_{(t^-,t]}$ between consecutive update times $t^-$ and $t$.
After each packet, we seek features $\hat{F}_t$ that match $F_t=\Phi(I_t)$, the features of a frame $I_t$ at time $t$.
Since this frame is unavailable at inference, $\hat{F}_t$ must be derived from the anchor features and the events alone.
ALIVE computes $\hat{F}_t$ recurrently (\cref{fig:method}) with an updater $\mathcal{U}_\theta$ that maintains a feature state $U_t$ of the same form as $F_0$ and a memory $h_t$ on the same token grid, initialized as $U_0=F_0$ and $h_0=0$ and updated with every packet as $(U_t, h_t, \hat{F}_t) = \mathcal{U}_\theta(U_{t^-}, h_{t^-}, \mathcal{E}_{(t^-,t]})$.
Each update reads only the new packet and a fixed-size state, and its cost therefore does not grow with the time since the anchor.
Moreover, the output lies in the VFM feature space: heads trained on VFM features can read it directly, and the VFM itself can supervise the update.

\subsection{ALIVE Updater}
\label{sec:updater}

We instantiate $\Phi$ with DINOv3 ViT-B/16~\cite{dinov3} and use the outputs of its 3rd, 6th, 9th, and 12th blocks ($L=4$), each with 768 channels on a token grid of stride 16.
As the camera and objects move, the same content appears at other positions, and its features should move with it.
ALIVE therefore first transports the features along the motion observed by the events and then corrects them where transport falls short (\cref{fig:method}a).
Spatial refinement further adjusts each output token using its neighbors.
All these steps are conditioned on a memory that accumulates the information of past event packets.

\paragraph{Event encoder.}
We accumulate the events of each packet in a voxel grid~\cite{zhu2019unsupervised} with five temporal bins per polarity.
A convolutional encoder, trained from scratch, maps the grid to features at stride 8.
Its architecture follows the RAFT feature encoder~\cite{raft}, which E-RAFT~\cite{eraft} applies to events.
A $2\times2$ pixel unshuffle places these features on the token grid, and a $1\times1$ convolution compresses them into a 64-channel event descriptor $e_t$.

\paragraph{Motion-guided transport.}
When content moves, the features of a token should come from the position that its content occupied before the packet.
From $e_t$, the memory $h_{t^-}$, and the two shallowest layers of $U_{t^-}$, two convolutional blocks predict a displacement $d_t(p)$ for every token $p$, bounded to two tokens per update.
The previous features of all layers and the previous memory are then sampled at the displaced positions with bilinear interpolation: $\tilde{U}_t^\ell(p) = U_{t^-}^\ell\big(p - d_t(p)\big)$ and $\tilde{h}_t(p) = h_{t^-}\big(p - d_t(p)\big)$.
All layers share one displacement, which keeps them aligned.
The memory moves with the same displacement and thus stays attached to the content it describes.

\paragraph{ConvGRU memory.}
A packet covers only a short interval, whereas a scene change, such as the motion of an object, extends over many packets.
The memory therefore accumulates event information across packets: after transport, a convolutional GRU~\cite{ballas2016delving} updates it with the new descriptor, $h_t = \mathrm{ConvGRU}(e_t, \tilde{h}_t)$.
The memory conditions residual correction and spatial refinement, and the next displacement prediction reads it as $h_{t^-}$.

\paragraph{Residual correction.}
The transported features can still deviate from the target, for example where the displacement is inaccurate or where the appearance of the content changes.
We therefore introduce residual correction, which predicts a residual $\Delta_t^\ell$ for every layer from $\tilde{U}_t^\ell$, $h_t$, and $e_t$ and adds it to the transported features, $U_t^\ell = \tilde{U}_t^\ell + \Delta_t^\ell$.
In this residual form, the correction only needs to account for what transport misses.
A ConvNeXt block~\cite{convnext} shared by all layers computes the residual in a 64-channel space.

\paragraph{Spatial refinement.}
Transport assigns one displacement to each token, which describes the motion well when all content within the token moves together.
At the boundary of a moving object, however, a token covers parts of the object and of its surroundings, which move differently, and bilinear sampling also mixes the features of both sides, whereas its neighbors often lie entirely on one side.
We therefore introduce spatial refinement, in which each token recombines the features of its $3\times3$ neighborhood $\mathcal{N}(p)$, including $p$ itself: $\hat{F}_t^\ell(p) = U_t^\ell(p) + \sum_{q\in\mathcal{N}(p)} k_t^\ell(p,q)\, U_t^\ell(q)$,
where the nine signed weights satisfy $\sum_{q} k_t^\ell(p,q)=0$ and $\sum_{q} |k_t^\ell(p,q)|\leq 1$.
Since they sum to zero, uniform regions remain unchanged, and the refinement acts only where neighboring features differ.
For example, a weight $k$ on a neighbor $q$ and $-k$ on $p$ give $\hat{F}_t^\ell(p) = U_t^\ell(p) + k\,[U_t^\ell(q) - U_t^\ell(p)]$, which brings the feature of $p$ closer to that of $q$ for $k>0$.
The $\ell_1$ bound limits each change to the largest difference between the token and its neighbors.
A small head predicts the weights from $h_t$, $U_t^\ell$, and the uncompressed event features.
Since all channels share these weights, the refinement only recombines existing features.
Applied only to the output, with the unrefined $U_t$ kept in the state, it refines each output exactly once.

\subsection{Task-Agnostic Training}
\label{sec:training}

We train ALIVE on synthetic sequences that provide frames at a high rate with synchronized events (\cref{sec:exp_setup}), which allows us to supervise every update with the VFM features of the frame at the end of its packet.
Starting from an anchor frame, ALIVE processes the packets of a sequence one by one, and the frozen VFM encodes the frame after every packet into the teacher features $F_t$.
All modules are trained jointly in a single stage, with gradients through all updates of a sequence and without task labels or heads.
We initialize the output layers of the displacement, residual, and refinement modules to zero, and training thus starts from the identity update.
Three losses compare $\hat{F}_t$ with $F_t$: per token, between tokens, and over time.
Below, $\mathbb{E}$ averages over update times, layers, and tokens or token pairs, and $\rho$ is the Huber loss~\cite{huber1964robust}, applied to features normalized per layer by their training-set root-mean-square value.

\paragraph{Tokenwise distillation.}
Each output token should match its teacher token in value and direction.
For two token features $x$ and $y$, we measure their discrepancy as $\delta(x,y) = \rho(x-y) + 1 - \cos(x,y)$ and define $\mathcal{L}_{\mathrm{distill}} = \mathbb{E}\,\delta(\hat{F}_t^\ell, F_t^\ell)$.

\paragraph{Spatial-token relation.}
Dense heads process each token together with its spatial context, but the tokenwise loss constrains each token on its own.
We therefore compare the cosine similarities $\hat{s}_t^\ell(p,q)$ and $s_t^\ell(p,q)$ between tokens $p$ and $q$ in the output and teacher features~\cite{structureddistill}, $\mathcal{L}_{\mathrm{spatial}} = \sum_{g} w_g\, \mathbb{E}_{(p,q)\in\mathcal{P}_g} |\hat{s}_t^\ell(p,q) - s_t^\ell(p,q)|$.
The set $\mathcal{P}_{\mathrm{near}}$ contains adjacent pairs, and $\mathcal{P}_{\mathrm{mid}}$ and $\mathcal{P}_{\mathrm{far}}$ contain pairs 4 and 12 tokens apart, weighted by 0.5, 0.25 and 0.25.

\paragraph{Temporal increment.}
The two losses above do not explicitly constrain how the features change between updates.
We therefore supervise this change directly with the temporal increment loss, $\mathcal{L}_{\mathrm{temporal}} = \mathbb{E}\, \rho\big((\hat{F}_t^\ell - \hat{F}_{t'}^\ell) - (F_t^\ell - F_{t'}^\ell)\big)$ for $t-t'$ of 5\,ms and 20\,ms.
The total training objective is $\mathcal{L} = \mathcal{L}_{\mathrm{distill}} + \mathcal{L}_{\mathrm{spatial}} + \mathcal{L}_{\mathrm{temporal}}$.

\subsection{Zero-Shot Transfer}
\label{sec:transfer}

After training, ALIVE transfers to real event streams without any adaptation (\cref{fig:method}b).
Frozen heads trained on VFM features of real RGB images read its output directly, and a single updater can serve different tasks and datasets.

\paragraph{Moment matching.}
Although feature supervision encourages ALIVE to reproduce VFM features, deviations that accumulate over many updates and differences between synthetic and real events can make the output channel statistics drift, which shifts the inputs of the frozen heads away from their training features.
The anchor features, extracted from a real frame of the same scene, provide a reference.
If the content in view only moved, the same feature values would appear at other tokens, and the distribution of each channel over the image would remain approximately the same.
We therefore introduce moment matching, which aligns the first two moments of every output channel, its mean and its variance, with those of the anchor, using the channel-wise affine transformation of adaptive instance normalization~\cite{adain}.
With the mean $\mu_t$ and variance $\sigma_t^2$ of an output channel over all tokens of the image, and the anchor moments $\mu_0$ and $\sigma_0^2$ computed from $F_0$ in the same way, the moment-matched output is $\bar{F}_t = (\sigma_0/\sigma_t)\odot(\hat{F}_t-\mu_t)+\mu_0$.
Every channel of $\bar{F}_t$ thus has the mean and variance of the anchor, while its standardized values, which form the spatial pattern of the channel, remain those predicted by ALIVE.
Moment matching has no parameters and acts only on the output, leaving the recurrence exactly as in training.

\section{Experiments}
\label{sec:experiments}
\begin{table*}[t]
\centering
\caption{
Comparison with baselines.
Detection and segmentation scores are in \%, and RMSE is in meters.
The $\Delta$ row gives the absolute gain for detection and segmentation and the relative change for depth.
The cost of one feature update is measured on one NVIDIA GH200 (FP32, batch size 1, $640\times640$ inputs).
GFLOPs count convolutions, matrix multiplications, and attention.
Latency is the mean time per update over 40 consecutive updates and 12 runs on three DSEC clips, including event preprocessing but excluding anchor encoding, history initialization, and task heads.
\textbf{Bold} and \underline{underlined} values indicate the best and second-best results, excluding Current RGB.
}
\label{tab:main_results}
\begingroup
\small
\setlength{\tabcolsep}{2.45pt}
\renewcommand{\arraystretch}{1.15}
\begin{tabular}{@{}lcccccccccccc@{}}
\toprule
\multirow{3}{*}{Method} & \multicolumn{3}{c}{Detection} & \multicolumn{2}{c}{Segmentation} & \multicolumn{4}{c}{Depth (0.1--30\,m)} & \multicolumn{3}{c}{Cost per update} \\
\cmidrule(lr){2-4}\cmidrule(lr){5-6}\cmidrule(lr){7-10}\cmidrule(l){11-13}
 & \multicolumn{3}{c}{DSEC~\cite{dsec}} & DSEC~\cite{dsec} & DDD17~\cite{ddd17} & \multicolumn{2}{c}{DSEC~\cite{dsec}} & \multicolumn{2}{c}{CoSEC~\cite{cosec}} & \multirow{2}{*}{\footnotesize\shortstack{Params\\(M)}} & \multirow{2}{*}{\footnotesize\shortstack{GFLOPs\\$\downarrow$}} & \multirow{2}{*}{\footnotesize\shortstack{Latency\\(ms) $\downarrow$}} \\
\cmidrule(lr){2-4}\cmidrule(lr){5-5}\cmidrule(lr){6-6}\cmidrule(lr){7-8}\cmidrule(lr){9-10}
 & mAP $\uparrow$ & AP50 $\uparrow$ & AP75 $\uparrow$ & mIoU $\uparrow$ & mIoU $\uparrow$ & AbsRel $\downarrow$ & RMSE $\downarrow$ & AbsRel $\downarrow$ & RMSE $\downarrow$ & & & \\
\midrule
Stale RGB & 27.92 & 61.10 & 22.06 & 57.16 & 66.85 & 0.145 & 4.341 & \underline{0.182} & \underline{4.303} & -- & -- & -- \\
\midrule
E-RAFT~\cite{eraft} & 51.51 & 84.17 & 53.93 & \underline{65.51} & 69.85 & 0.182 & 4.418 & 0.195 & 4.335 & 5.33 & 690.1 & 71.56 \\
EDCFlow~\cite{edcflow} & \underline{51.93} & \underline{84.57} & \underline{54.48} & 65.40 & \underline{70.18} & 0.182 & 4.429 & 0.195 & 4.336 & 2.50 & 667.6 & 124.43 \\
\addlinespace[2.5pt]
E2VID~\cite{e2vid} & 33.48 & 61.90 & 31.98 & 48.63 & 25.38 & 0.255 & 6.403 & 0.256 & 5.555 & 10.71 & 741.5 & 28.96 \\
HyperE2VID~\cite{hypere2vid} & 44.67 & 77.45 & 44.38 & 55.10 & 43.79 & 0.190 & 4.927 & 0.299 & 6.239 & 10.15 & 714.8 & 29.62 \\
\addlinespace[2.5pt]
VFPSIE~\cite{vfpsie} & 48.44 & 81.45 & 48.81 & 60.60 & 69.41 & 0.144 & 3.987 & 0.215 & 4.514 & 2.15 & 422.7 & 20.47 \\
DI~\cite{brandli2014decompression} & 31.19 & 64.81 & 25.86 & 57.25 & 66.28 & 0.140 & 4.028 & 0.196 & 4.395 & 0.00 & \underline{369.5} & \underline{12.05} \\
\addlinespace[2.5pt]
ScaleEvent~\cite{scaleevent} & 32.67 & 61.32 & 31.03 & 48.29 & 20.14 & 0.225 & 6.054 & 0.295 & 6.723 & 85.67 & \underline{369.5} & 16.04 \\
\addlinespace[2.5pt]
LiFR-v2~\cite{lifrv2} & 36.45 & 69.24 & 35.09 & 57.37 & 56.81 & \underline{0.137} & \underline{3.849} & 0.234 & 5.303 & 123.30 & 687.7 & 40.35 \\
\addlinespace[2.5pt]
\rowcolor[HTML]{E6F2F3}
ALIVE (Ours) & \textbf{52.55} & \textbf{85.02} & \textbf{56.27} & \textbf{66.30} & \textbf{72.31} & \textbf{0.129} & \textbf{3.741} & \textbf{0.173} & \textbf{4.011} & 2.50 & \textbf{65.0} & \textbf{4.66} \\
\rowcolor[HTML]{E6F2F3}
{\footnotesize $\Delta$ vs.\ Stale} & {\footnotesize \textcolor[HTML]{2E7D32}{+24.63}} & {\footnotesize \textcolor[HTML]{2E7D32}{+23.92}} & {\footnotesize \textcolor[HTML]{2E7D32}{+34.21}} & {\footnotesize \textcolor[HTML]{2E7D32}{+9.14}} & {\footnotesize \textcolor[HTML]{2E7D32}{+5.46}} & {\footnotesize \textcolor[HTML]{2E7D32}{$-$11.0\%}} & {\footnotesize \textcolor[HTML]{2E7D32}{$-$13.8\%}} & {\footnotesize \textcolor[HTML]{2E7D32}{$-$4.9\%}} & {\footnotesize \textcolor[HTML]{2E7D32}{$-$6.8\%}} & & & \\
\midrule
Current RGB & 64.43 & 88.85 & 74.57 & 73.32 & 76.94 & 0.096 & 2.852 & 0.166 & 3.813 & 85.67 & 369.5 & 11.80 \\
\bottomrule
\end{tabular}
\endgroup
\end{table*}

\subsection{Experimental Setup}
\label{sec:exp_setup}

\paragraph{Training data.}
We train ALIVE on TartanAir V2~\cite{tartanair,tartanairv2}, which provides high-frame-rate synthetic video with events.
Its 42 training trajectories yield 20,183 sequences of 200\,ms after an anchor frame, each with 40 packets of 5\,ms and a teacher frame after every packet.
We train for 3,240 steps with a batch size of 64.

\paragraph{Datasets and tasks.}
We evaluate on three real-world datasets recorded with different event cameras.
DSEC~\cite{dsec} pairs RGB frames at 20\,Hz with a Prophesee Gen3.1 event camera ($640\times480$).
We use it for detection with the labels of DSEC-Detection~\cite{gehrig2024lowlatency}, for segmentation with the 11 classes of ESS~\cite{ess}, and for depth with its 10\,Hz disparity ground truth.
DDD17~\cite{ddd17} records grayscale frames about every 29\,ms and events with one DAVIS346B sensor ($346\times260$), which we use for segmentation with the 6 classes of EV-SegNet~\cite{alonso2019evsegnet}.
CoSEC~\cite{cosec} aligns RGB frames with a Prophesee EVK4 event camera ($1280\times720$) and provides LiDAR depth for its frames.
Detection and segmentation are evaluated on the official test splits, and depth on six held-out DSEC recordings and the 14 Village trajectories of CoSEC.

\paragraph{Task heads and protocol.}
For each setting, a head trained on DINOv3 features of the training frames is frozen: FPN~\cite{fpn} with Faster R-CNN~\cite{fasterrcnn} for detection, UPerNet~\cite{upernet} for segmentation, and a DPT decoder~\cite{dpt} for metric depth.
All methods share these heads and use one checkpoint each for all settings, which isolates the quality of their features.
Each test sample pairs a labeled target frame with an earlier anchor frame, and each method receives the anchor and the events in between.
We call the time between them the anchor age.
To compare ages on the same content, we fix the targets and use each of their four (DSEC) or six (DDD17) preceding frames as anchors, with ages of 50--200\,ms and 29--176\,ms.
This gives 36,260, 10,936, 12,216, and 23,340 samples for DSEC detection, segmentation, depth, and DDD17.
As the irregular frames of CoSEC do not allow common ages, we pool its 1,893 consecutive frame pairs (31\,ms--1\,s), and \cref{fig:accuracy_efficiency} covers the other settings.

\paragraph{Metrics.}
We report mAP, AP50, and AP75 on vehicles and pedestrians~\cite{gehrig2024lowlatency,flexevent}, mIoU, and AbsRel and RMSE for metric depth within 0.1--30\,m without scale alignment, pooled over all anchor ages unless stated otherwise.

\paragraph{Baselines.}
We compare with representative methods of each approach in \cref{sec:related}, all yielding DINOv3 features for the same heads: flow warping (E-RAFT~\cite{eraft}, EDCFlow~\cite{edcflow}), image reconstruction from events (E2VID~\cite{e2vid}, HyperE2VID~\cite{hypere2vid}) or from the anchor and events (VFPSIE~\cite{vfpsie}, direct integration (DI)~\cite{brandli2014decompression}) followed by DINOv3, event-to-feature mapping (ScaleEvent~\cite{scaleevent}), and feature propagation (LiFR-v2~\cite{lifrv2}).
All learned baselines except E2VID and HyperE2VID (released weights) are trained or fine-tuned on TartanAir V2.
EDCFlow and LiFR-v2 also start from components trained on DSEC, and DI fits its contrast thresholds on each training split, whereas ALIVE never sees a real recording.
Stale RGB reads the anchor features, and Current RGB encodes the target frame itself, an upper bound unavailable between frames.

\subsection{Comparison with Baselines}
\label{sec:comparison}
\begin{figure*}[t]
    \centering
    \includegraphics[width=\textwidth]{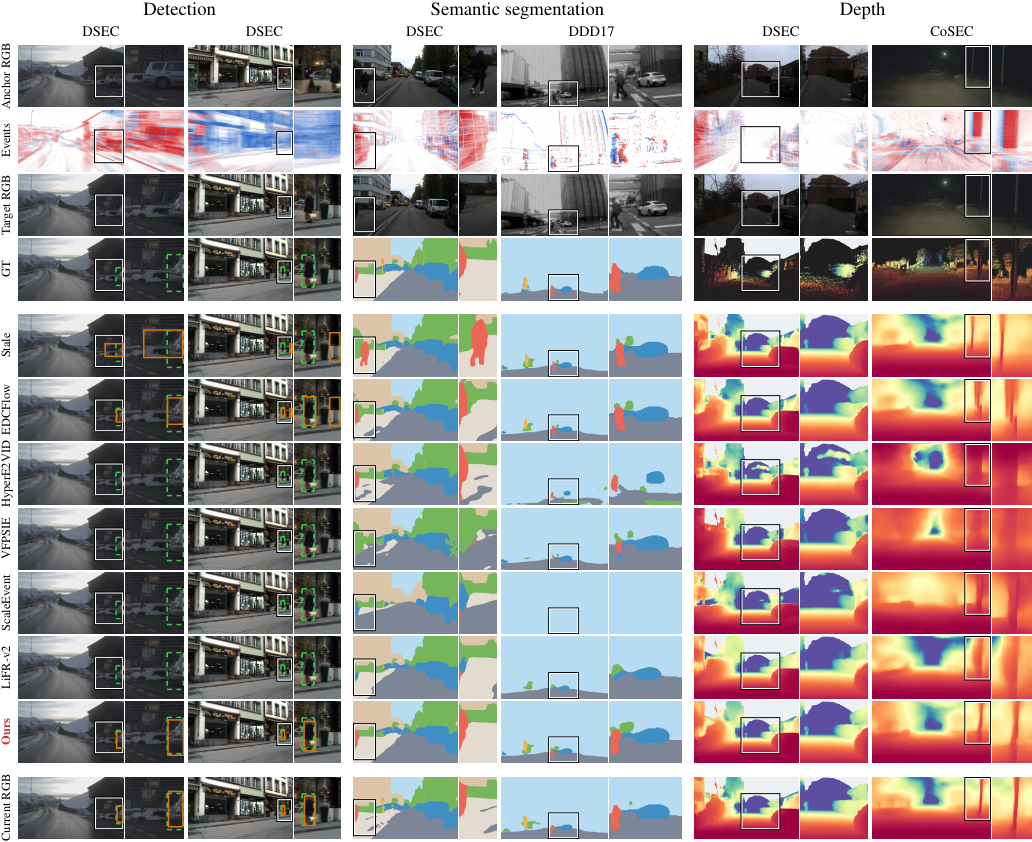}
    \caption{
        \textbf{Qualitative comparison on test samples.}
        Each column shows one sample at its target time, and each view is followed by a zoom of the framed region.
        The top four rows show the anchor frame, the events between the anchor and the target time (red: positive, blue: negative), the target frame, and the ground truth.
        The other rows show the predictions of the frozen heads from the features of each method.
        Detection: predictions in orange and ground truth dashed in green.
        Depth: red (near) to blue (far), with one range per sample.
        Apart from Current RGB, no method receives the target frame, and the ground truth is shown for reference only.
    }
    \label{fig:qualitative}
\end{figure*}

As \cref{tab:main_results} shows, ALIVE is the most accurate method in every metric of all five settings.
It achieves this with a single checkpoint that has never seen a real recording, on three event cameras that differ from each other and from the simulated training events.
Compared with the stale features, ALIVE gains 24.6~mAP in DSEC detection and 34.2 in AP75, the metric that requires the most precise localization.
Our method also improves segmentation by 9.1 and 5.5~mIoU on DSEC and DDD17 and reduces AbsRel by 11.0\% on DSEC and 4.9\% on CoSEC, where even Current RGB improves on the stale features by only 9\% (34\% on DSEC).

The baselines show what such an update requires.
First, the anchor frame is essential: the methods that ignore it fall below the stale features in all segmentation and depth settings, since events mainly record moving edges and little of the static scene layout.
Second, the anchor frame alone is not sufficient: flow warping yields the strongest baselines for detection and segmentation, which shows that motion accounts for much of the change between frames, but it only relocates the features and degrades depth below the stale features.
LiFR-v2 is the strongest baseline for depth on DSEC but trails flow warping by 15~mAP in detection.
Updating the anchor image keeps its appearance, but VFPSIE stays below flow warping in detection and segmentation, and DI stays close to the stale features.
Only ALIVE, which transports the anchor features and corrects what transport misses, is the most accurate in all settings.

\cref{fig:qualitative} shows examples.
The stale features keep objects at their anchor positions, which causes missed objects and false positives, as for the car and the pedestrian in the detection examples.
Flow warping moves objects but leaves copies behind, such as the second box of the pedestrian and the split lamp post in the CoSEC example.
Where events are sparse, as on the distant building in the DSEC depth example, the event-only methods show artifacts, and small moving objects are often lost.
ALIVE places all of these objects at their target positions, close to Current RGB.

\subsection{Accuracy over Anchor Age}
\label{sec:age}

As the bottom row of \cref{fig:accuracy_efficiency} shows, the stale features lose accuracy quickly as objects move away from their anchor positions.
ALIVE loses accuracy much more slowly, and its gain grows with the anchor age in all four settings, up to 30.9~mAP and 11.6~mIoU on DSEC, 7.6~mIoU on DDD17, and 15.9\% in DSEC AbsRel.
The strongest baseline changes with setting and age, but ALIVE remains the most accurate method at every age.

\subsection{Efficiency}
\label{sec:efficiency}

ALIVE needs 65.0~GFLOPs and 4.66\,ms per update with 2.50M parameters (\cref{tab:main_results}), about 3\% of DINOv3-B, because it updates the existing features instead of running the VFM again.
Every baseline costs at least as much as encoding a frame with DINOv3 (369.5~GFLOPs).
Compared with re-encoding the current frame, ALIVE reduces the computation by 82.4\% and the latency by a factor of 2.5, and it is the only method that processes each 5\,ms packet within 5\,ms and can thus update the features at 200\,Hz.
In all four settings, ALIVE is both the most accurate and the fastest method (\cref{fig:accuracy_efficiency}, top).

\subsection{Ablation Study}
\label{sec:ablation}
\begin{table}[t]
\centering
\caption{
Ablation on DSEC of the inputs and modules (top) and of the training losses (bottom).
Removing the memory also removes memory transport.
\textbf{Bold} and \underline{underlined} values indicate the best and second-best results.
}
\label{tab:ablation_dsec}
\begingroup
\small
\setlength{\tabcolsep}{3.2pt}
\renewcommand{\arraystretch}{1.15}
\begin{tabular}{@{}lccccc@{}}
\toprule
\multirow{2}{*}{Variant}
& \multicolumn{2}{c}{Det.\ $\uparrow$}
& Seg.\ $\uparrow$
& \multicolumn{2}{c}{Depth $\downarrow$} \\
\cmidrule(lr){2-3}\cmidrule(lr){4-4}\cmidrule(lr){5-6}
& mAP & AP75 & mIoU & AbsRel & RMSE \\
\midrule
\rowcolor[HTML]{E6F2F3}
Full model & \textbf{52.55} & \textbf{56.27} & \textbf{66.30} & \underline{0.129} & \underline{3.741} \\
w/o events            & 27.80 & 21.35 & 56.51 & 0.160 & 4.641 \\
w/o feature transport & 42.10 & 40.43 & 61.02 & 0.140 & 4.078 \\
w/o memory transport  & 51.56 & 54.13 & 65.86 & 0.132 & 3.788 \\
w/o memory            & 50.19 & 51.52 & 65.07 & 0.134 & 3.897 \\
w/o residual correction & 51.80 & 54.97 & 65.98 & 0.133 & 3.810 \\
w/o spatial refinement  & 51.74 & 54.38 & 65.57 & \textbf{0.127} & \textbf{3.712} \\
\begin{tabular}[c]{@{}l@{}}w/o residual correction\\\phantom{w/o }\& spatial refinement\end{tabular} & 49.76 & 51.37 & 64.99 & 0.134 & 3.877 \\
w/o moment matching   & 52.02 & 54.95 & 65.97 & 0.162 & 4.361 \\
\midrule
w/o $\mathcal{L}_{\mathrm{spatial}}$  & \underline{52.50} & 55.40 & 66.11 & 0.139 & 3.997 \\
w/o $\mathcal{L}_{\mathrm{temporal}}$ & 52.41 & \underline{55.93} & \underline{66.27} & 0.130 & 3.766 \\
$\mathcal{L}_{\mathrm{distill}}$ only & 52.38 & 55.24 & 66.04 & 0.138 & 4.000 \\
\bottomrule
\end{tabular}
\endgroup
\end{table}

We ablate ALIVE on DSEC (\cref{tab:ablation_dsec}), retraining every variant except the one without moment matching, which only removes this test-time step.

\paragraph{Modules and inputs.}
Events are indispensable.
They are the only source of information about the scene after the anchor, and without them, the updater keeps all its modules but falls to the accuracy of the stale features or below.
Feature transport is the most important module, in line with the strength of flow warping: without it, detection drops by 10.5~mAP and segmentation by 5.3~mIoU.
The memory adds 2.4~mAP, of which 1.0 is lost when it does not move with the features.
Residual correction and spatial refinement correct different errors: spatial refinement, which acts where neighboring features differ, matters more for AP75 and mIoU, whereas residual correction matters more for depth.
Together, they give the best detection and segmentation results.
Among the modules, moment matching has by far the largest effect on depth (AbsRel 0.162 $\rightarrow$ 0.129), which indicates that metric depth depends strongly on the channel statistics of the features.

\paragraph{Losses.}
Trained with $\mathcal{L}_{\mathrm{distill}}$ alone, ALIVE already outperforms all baselines in detection and segmentation on DSEC (\cref{tab:main_results,tab:ablation_dsec}), which shows that most of its accuracy stems from our architecture.
$\mathcal{L}_{\mathrm{spatial}}$, which constrains the relations between tokens, matters most for depth and AP75, which both depend on the spatial structure of the features.
$\mathcal{L}_{\mathrm{temporal}}$, which constrains the change between updates, improves every metric by smaller margins, except for mAP.
The two losses are complementary, and combining them gives the best result in every metric.

\section{Conclusion}
\label{sec:conclusion}

We presented ALIVE, a lightweight event-driven updater that keeps VFM features alive through feature transport, residual correction, spatial refinement, and moment matching. Trained on synthetic data without task labels, ALIVE generalizes to three real-world datasets, outperforming all baselines across five settings while supporting 200\,Hz updates with 82.4\% fewer FLOPs than frame re-encoding.

\paragraph{Acknowledgments.}
This work was supported by the European Union's Horizon Europe Research and Innovation Programme under grant agreement No.~101120732 (AUTOASSESS) and the European Research Council (ERC) under grant agreement No.~864042 (AGILEFLIGHT).

{
    \small
    \bibliographystyle{ieeenat_fullname}
    \bibliography{main}
}

\end{document}